\documentclass[10pt]{article}

\usepackage[preprint]{tmlr}       
\usepackage{booktabs}
\usepackage{amsmath}
\usepackage{amssymb}
\usepackage{graphicx}
\graphicspath{{figures/}}
\usepackage{microtype}
\usepackage{hyperref}
\usepackage{url}

\title{How Faithful Is Attribution for Sales Forecasting?\\
A Counterfactual Study}

\author{\name Glib Kechyn}

\begin{document}
\maketitle

\begin{abstract}
Deep models for sales forecasting, such as WaveNet-style dilated convolutional
networks, are accurate but opaque: when a single model predicts sales for one of
many series, it offers no account of \emph{why}. We add a post-hoc,
architecture-agnostic counterfactual interpretability layer to a
multi-series WaveNet forecaster trained on the full Corporaci\'on Favorita
grocery dataset (174{,}685 series over 1{,}688 days). The method decomposes
each forecast into contributions that sum \emph{exactly} to the predicted
value, avoiding the allocation artifacts we observed with additive
SHAP-style attribution. We evaluate faithfulness with a deletion/insertion
protocol and find a statistically significant effect on both tests (deletion gap
$0.22$, $p<0.001$; insertion gap $0.27$, $p<0.01$; robust across five
background-sampling seeds), establishing that the
attributions reflect genuine model behavior rather than plausible-looking
artifacts. We then characterize, honestly, where attribution is and is not
informative: reliance on the promotion signal is heterogeneous across series
(median ratio ${\approx}1.0$, with roughly $20\%$ of series showing a strong
effect), and the model captures the \emph{shape} of the weekly sales cycle
(day-of-week $r{=}0.78$) while systematically under-predicting its
amplitude. Our contribution is not improved accuracy but an interpretability
layer with a rigorous faithfulness evaluation and a candid account of its
limits.
\end{abstract}

\section{Introduction}
\label{sec:intro}
A sales forecast is not a passive prediction; it is acted upon. Grocery
retailers use per-item forecasts to decide how much to order, and the cost of
error is concrete and asymmetric: over-order and perishable stock is wasted,
under-order and shelves empty and customers leave \citep{kechyn2018sales}. When
a forecast drives such a decision, the practitioner's next question is rarely
just \emph{how much?} but \emph{why?}---is this number high because the item is
trending, because a promotion is running, or because of a recurring weekly
pattern? A model that answers only the first question leaves the decision-maker
to trust or distrust it wholesale.

Modern multi-series forecasters answer the first question well and the second
poorly. A single deep model trained across a large panel of related series
attains strong accuracy \citep{kechyn2018sales}, but its output is one opaque
number. Architectures that build in interpretability exist---N-BEATS
\citep{oreshkin2020nbeats} and the Temporal Fusion Transformer
\citep{lim2021temporal} are prominent examples---but they obtain it by design,
which requires adopting and retraining into that specific architecture. For a
forecaster that is already trained and deployed, that is a steep price merely to
ask why it produced yesterday's number.

We take the complementary, post-hoc route: we explain an existing forecaster
without changing it. Two ideas do the work. \emph{Attribution} asks which parts
of the input are responsible for a given prediction, decomposing a single
forecast into the credit due to, say, the item's baseline level, its recent
trend, and an active promotion. \emph{Counterfactual} attribution obtains that
decomposition by asking, for each part of the input, what the model would
predict if that part were withheld and replaced by a neutral baseline; the
resulting change in the prediction is that part's contribution. Because each
contribution is a real, measured difference between model predictions rather
than an estimated share of a fixed total, the contributions sum \emph{exactly}
to the forecast being explained.

An explanation is useful only if it is \emph{faithful}---if it reflects what the
model actually computes rather than merely looking plausible. This is our
central concern. Using deletion and insertion tests \citep{petsiuk2018rise}, we
show that the attributions are faithful with high statistical significance
(deletion $p<0.001$, insertion $p<0.01$), a form of validation that, to our knowledge, remains far less
common in multi-series forecasting than in the vision settings where deletion
and insertion tests originated. We are equally deliberate about the method's
limits: reliance on the promotion signal is \emph{heterogeneous} across
series---pronounced for a minority, negligible for the majority---and the model
captures the shape of the weekly cycle while damping its amplitude. We report
these as findings, not failures: an interpretability method earns trust by
showing where its explanations are informative and where they are not.

We demonstrate the method on the WaveNet-style forecaster of
\citet{kechyn2018sales} and the full Corporaci\'on Favorita grocery dataset, but
the attribution construction treats the model as a black box and is not specific
to that architecture (Section~\ref{sec:related}). Our contributions are:
\begin{itemize}
  \item A post-hoc, architecture-agnostic counterfactual attribution method for
  multi-series forecasters, whose contributions sum exactly to the predicted
  value (Section~\ref{sec:method}).
  \item A faithfulness evaluation via deletion and insertion tests, showing that
  the attributions reflect genuine model behavior at high significance
  (Section~\ref{sec:faithfulness}).
  \item An honest characterization of when attribution is informative: a
  heterogeneity analysis of promotion reliance and an analysis of weekly-cycle
  capture, including where the model falls short (Section~\ref{sec:attribution}).
\end{itemize}

We deliberately make no claim of improved forecast accuracy. Our
covariate-augmented model modestly improves on a sales-only baseline
(NWRMSLE $0.6137 \rightarrow 0.6102$), reported for completeness; the
contribution is interpretability that can be trusted, not a new accuracy record.

\section{Related Work}
\label{sec:related}
\paragraph{WaveNet and its use in forecasting.} The dilated causal convolution
with gated activations originates with WaveNet \citep{vandenoord2016wavenet}, a
generative model for raw audio. \citet{kechyn2018sales} adapted the architecture
to retail sales forecasting on the Corporaci\'on Favorita dataset, training a
single convolutional model across a large panel of series and reaching a strong
competition result. That work is an applied, competition-oriented account: it
established the forecaster but reported no interpretability layer and no
evaluation of whether any explanation of its predictions would be faithful. Our
model follows its shared-panel setup (Section~\ref{sec:model}); our contribution
is the attribution layer and the faithfulness evaluation it lacked.

\paragraph{Interpretability in modern forecasters.} The field has since produced
architectures that treat interpretability as a design goal rather than an
afterthought. N-BEATS \citep{oreshkin2020nbeats} builds an explicit basis
expansion into the network, so that trend and seasonality emerge as separable,
inspectable components. The Temporal Fusion Transformer
\citep{lim2021temporal} goes further, embedding variable-selection networks and
interpretable attention heads directly into the model, so that variable
importance and temporal focus can be read off the trained network. PatchTST
\citep{nie2023patchtst} tokenizes series into patches for long-horizon
Transformer forecasting; interpretability is not its aim, but it exemplifies the
direction the architecture literature has taken.

\paragraph{Where this work sits.} These interpretable-by-design models share a
premise: to obtain interpretability, adopt their architecture. That is a real
cost when a forecaster is already trained and deployed---switching to TFT to
learn why yesterday's forecast came out as it did means retraining, revalidating,
and re-deploying a different model. Our approach is complementary and
post-hoc: it explains an existing forecaster without modifying or retraining it.
The attribution method treats the model as a black box, requiring only that it
map an input window and a series identifier to a prediction; nothing in the
construction is specific to WaveNet, and it applies unchanged to any forecaster
matching that signature. We demonstrate it on a WaveNet-style model because that
is the forecaster of \citet{kechyn2018sales}, not because the method depends on
it.

\paragraph{Faithfulness and attribution.} Post-hoc attribution has a large
literature in classification, including additive feature-attribution methods
such as SHAP \citep{lundberg2017shap}. A recurring concern is
\emph{faithfulness}: an attribution can look reasonable while failing to reflect
what the model actually computes. Deletion and insertion tests
\citep{petsiuk2018rise} probe this by perturbing inputs in attribution order and
measuring the effect on the prediction. Such evaluation is well established in
vision but, to our knowledge, is applied less often in multi-series forecasting,
where interpretability work more commonly reports attributions than tests their
faithfulness. We adopt the
deletion/insertion protocol as our central evaluation (Section~\ref{sec:faithfulness}).

\section{Model}
\label{sec:model}
\paragraph{Setup.} We forecast each (store, item) series individually but train
one model across the whole panel. For a series $s$ and day $t$, the model maps a
window of the $L$ most recent days to the next $H$ days of sales.

\paragraph{Architecture.} The backbone is a stack of dilated causal
convolutions in the style of WaveNet \citep{vandenoord2016wavenet}. Each block
uses the gated activation $\tanh(W_f * x) \odot \sigma(W_g * x)$, where $*$ is a
dilated convolution and $\odot$ the elementwise product. We stack eight blocks
with dilations $1, 2, 4, \dots, 128$, $32$ filters per layer, and kernel size
$3$; the resulting receptive field spans the full input window. Causality is
enforced by left-only zero padding of width $d(k-1)$ for dilation $d$ and kernel
size $k$, so no output position ever depends on a future input and there is no
leakage. Per-block skip connections are summed across the stack; the aggregate
is passed through two $1{\times}1$ convolutions, averaged over the time axis, and
mapped by a linear layer to the $H$-step forecast.

\paragraph{Multi-series conditioning.} A single set of weights is shared across
all $174{,}685$ series. To let those shared weights serve heterogeneous series,
each series carries a learned embedding $e_s \in \mathbb{R}^{8}$ supplied
alongside the input signal. The network input has shape $(\text{batch}, 1+C, L)$,
where channel $0$ is sales and the remaining $C$ channels carry covariates; here
$C = 1$ (onpromotion). We use input length $L = 90$ and horizon $H = 16$.

\paragraph{Objective and scaling.} We train with an NWRMSLE-aligned objective in
$\log(1{+}x)$ space, applied to the \emph{sales channel only}; the binary
promotion channel is never transformed. Training on raw sales was unstable
across the panel's wide range of scales; the log transform aligns the objective
with the log-scale evaluation metric and compresses cross-series scale
differences.

\paragraph{Relation to prior WaveNet forecasters.} \citet{kechyn2018sales}
generated the 16-day horizon with a sequence-to-sequence encoder/decoder whose
encode and decode phases did not share parameters, letting the decoder absorb
error accumulation over the horizon. Our model instead emits the full horizon
directly from the dilated stack. The shared-across-the-panel design---one model
conditioned per series---follows their setup; the horizon mechanism does not.

\paragraph{Preprocessing.} We follow the data assumptions of
\citet{kechyn2018sales}. Missing (store, item)$\times$date combinations are
filled with zero sales; negative unit sales (returns) are clipped to zero;
onpromotion is mapped True/False to $1/0$, with missing values set to $0$. For
every series we hold out the final $16$ days for validation.

\section{Forecasting Results}
\label{sec:forecasting}
\paragraph{Metric.} We report NWRMSLE, the official competition metric: a
normalized weighted root mean squared logarithmic error in which perishable
items receive weight $1.25$ and all others $1.0$. Lower is better.

\paragraph{Framing.} We do not attempt to reproduce the competition leaderboard
result of \citet{kechyn2018sales}, which relied on an ensemble of five models
with exponential moving averaging and supplied the promotion signal as a
\emph{known-future} covariate. Our goal is different: a single, faithful
re-implementation in the same spirit, on the same data and metric, providing a
clean baseline for the interpretability contribution that follows. We therefore
report matched single-model comparisons rather than a leaderboard figure.

\paragraph{Effect of the promotion covariate.} Adding onpromotion improves
accuracy in the expected direction: a sales-only model reaches NWRMSLE $0.6137$,
and adding the covariate reaches $0.6102$. Both models are trained for $60$
epochs under identical settings and both converge, so the comparison is fair and
matched. The improvement is real but modest. We use onpromotion because it is
the covariate validated by \citet{kechyn2018sales}, who tested oil prices,
holidays, and transactions and discarded them as uncorrelated with the target.

\paragraph{Past-only promotion.} We deliberately supply onpromotion as a
\emph{past-only} signal, in contrast to \citet{kechyn2018sales}, who
additionally shifted it into the future. Restricting the model to observed, past
promotion status keeps the setup aligned with a deployment in which future
promotion schedules may be unknown at inference time; incorporating known-future
covariates is a natural extension we leave to future work.

\begin{figure}[t]
\centering
\includegraphics[width=\textwidth]{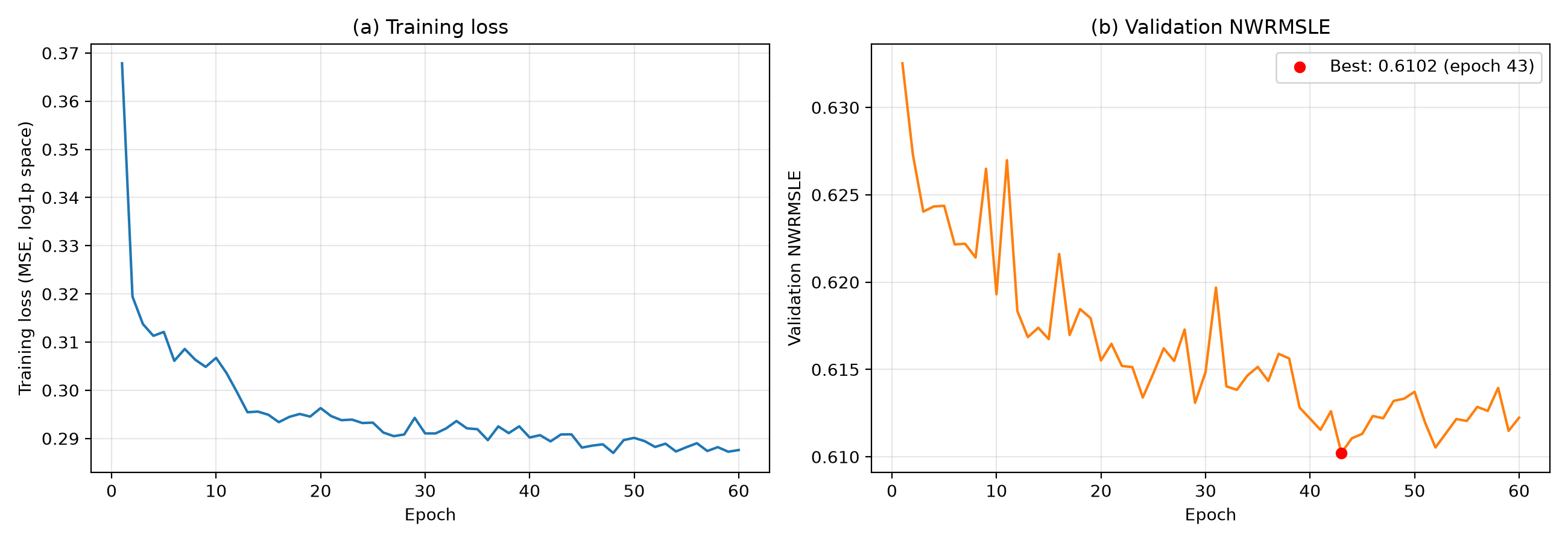}
\caption{Training loss (MSE in $\log(1{+}x)$ space, left) and validation NWRMSLE
(right) over $60$ epochs for the multi-series WaveNet with the onpromotion
covariate. Validation reaches its best value of $0.6102$ at epoch $43$; the
subsequent gentle decline in training loss alongside a slight rise in validation
indicates mild late overfitting.}
\label{fig:convergence}
\end{figure}

\paragraph{Convergence.} Training costs roughly $18$ seconds per epoch.
Validation NWRMSLE reaches its best value of $0.6102$ at epoch $43$ and stays
close to $0.61$ through epoch $60$; we select the epoch-$43$ checkpoint. After
that point the training loss continues to fall gently while validation drifts
upward only marginally (Figure~\ref{fig:convergence}), indicating at most mild
late-stage overfitting rather than a clean plateau.

\section{Interpretability Method}
\label{sec:method}
We attribute a forecast to interpretable groups of input positions through a
sequential counterfactual construction. Starting from a fully baselined
input---every channel set to a reference value---we reveal groups of true input
positions cumulatively, in a fixed order, and record the model's actual
prediction after each reveal. The contribution of a group is the change in the
measured prediction caused by revealing it.

Formally, let $g_1, \dots, g_K$ be an ordered partition of the input positions
into named groups (for example: seasonal pattern, recent trend, promotion).
Writing $f(\cdot)$ for the model and $m_k$ for the cumulative mask after
revealing $g_1, \dots, g_k$, the contribution of group $g_k$ is
$f(m_k) - f(m_{k-1})$, with $f(m_0)$ the fully baselined prediction. These
contributions telescope, so the baseline prediction plus the sum of all
contributions equals the true, fully revealed forecast \emph{exactly} (up to
floating-point error): there is no residual to allocate and no possibility of a
group's share being estimated away.

\paragraph{Why not additive allocation.} A natural alternative computes
per-position attributions (for example with a gradient-based SHAP estimator) and
sums them within each group. We found this collapses to near-zero for groups
containing very few positions, or positions whose signed attributions cancel:
for an item that is almost always on promotion, the promotion bucket can receive
an attributed share of essentially zero, falsely implying the model ignores
promotions. The counterfactual construction avoids this by measuring what the
model actually does when a group is withheld, rather than dividing a fixed
attribution total. We retain the gradient-based SHAP estimator only for the
per-timestep attribution visualizations and the ratio analysis of
Section~\ref{sec:attribution}, where a signed per-position signal is what we
want.

\paragraph{Scope.} These attributions are counterfactual statements about the
\emph{model}: they describe how its output responds to withholding parts of its
input. They are not claims about causal structure in the data-generating
process. We do not attempt causal discovery, and a large attribution to the
promotion channel means the model relies on that channel, not that promotions
cause sales in the world.

\paragraph{Code availability.} The method is implemented in an open-source
Python package.\footnote{Available at \url{https://github.com/YOUR-USERNAME/wavexplain}
and on PyPI: \texttt{pip install wavexplain==0.1.0}. Replace the URL with your
actual repository before posting.} All experiments in this paper were produced
with that package on the Corporaci\'on Favorita
dataset.\footnote{The dataset is the public Corporaci\'on Favorita Grocery Sales
Forecasting competition on Kaggle.}

\section{Faithfulness Evaluation}
\label{sec:faithfulness}
An attribution method can produce plausible-looking explanations that do not
reflect the model's actual behavior. We test faithfulness with the standard
deletion and insertion protocol \citep{petsiuk2018rise}, adapted to forecasting.

\paragraph{Protocol.} For each series we rank input positions by attributed
importance. In the \emph{deletion} test we mask positions from most to least
important and track the magnitude of change in the forecast; a faithful ranking
moves the forecast quickly, so the deletion curve should rise faster than one
produced by masking in random order. In the \emph{insertion} test we instead
reveal positions from most to least important, starting from a baselined input;
a faithful ranking recovers the original forecast quickly, so its curve should
fall toward zero faster than random. We summarize each test by the gap between
the attribution-ordered and random-ordered curves, and assess significance with
a one-sample $t$-test on the per-series gaps against zero. The evaluated series
are a uniform random sample drawn from the full panel of $174{,}685$ series
(fixed seed for reproducibility), with no filtering on sales volume or promotion
activity, so the evaluation is representative of the panel rather than of any
easily explained subset.

\begin{figure}[t]
\centering
\includegraphics[width=\textwidth]{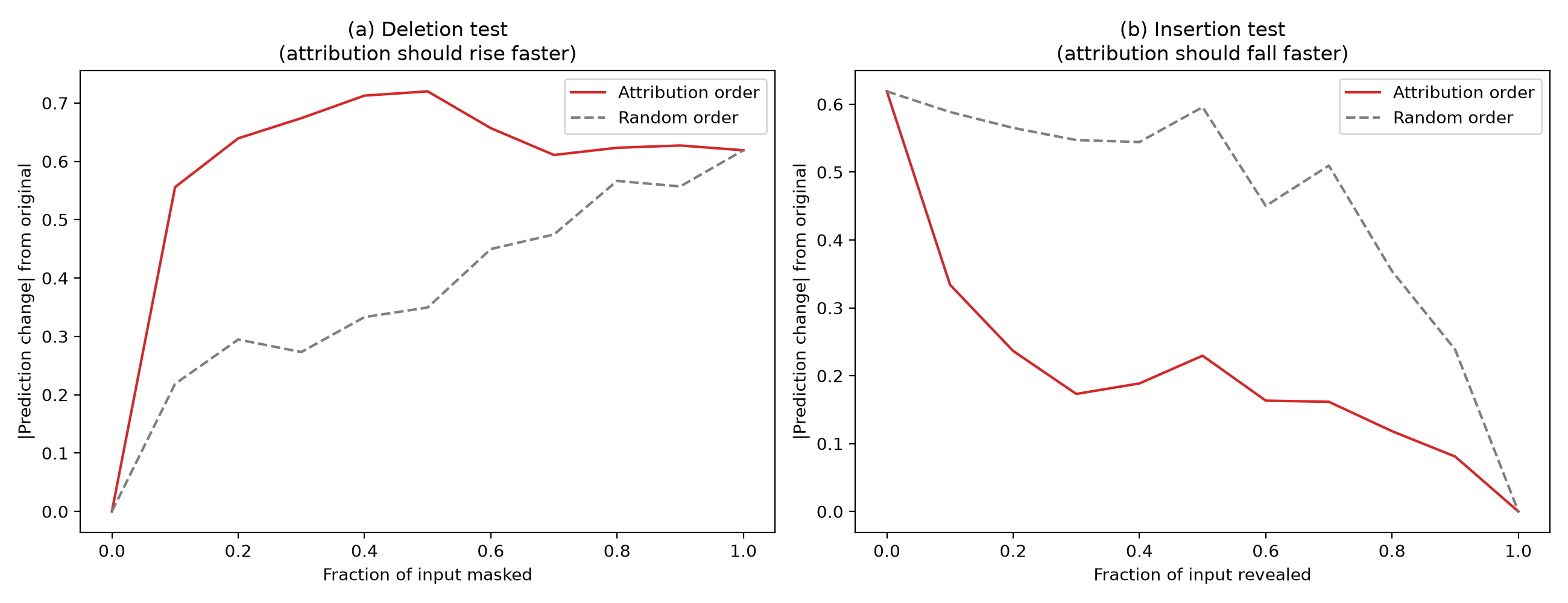}
\caption{Faithfulness of the counterfactual attributions, aggregated over $30$
series. Left: deletion---masking positions in attribution order (solid) degrades
the forecast far faster than random order (dashed). Right: insertion---revealing
positions in attribution order recovers the forecast far faster than random.
Results shown for background seed $0$; deletion is significant at $p<0.001$ and
insertion at $p<0.01$, on this and all five seeds tested.}
\label{fig:faithfulness}
\end{figure}

\paragraph{Results.} Across $30$ series the attribution ordering is decisively
more faithful than random on both tests (Figure~\ref{fig:faithfulness}). Because
the SHAP background is sampled at random, we fix its seed for reproducibility and
report the canonical run (seed $0$): the mean deletion gap is $0.217$
($t = 4.95$, $p = 2.9\times10^{-5}$) and the mean insertion gap is $0.271$
($t = 3.29$, $p = 2.7\times10^{-3}$); deletion is significant at $p<0.001$ and
insertion at $p<0.01$. To confirm the finding does not depend on the background
draw, we repeated the evaluation over five seeds: the deletion gap ranged
$0.216$--$0.263$ ($p$ from $2.9\times10^{-5}$ to $1.1\times10^{-4}$) and the
insertion gap $0.269$--$0.284$ ($p$ from $2.0\times10^{-3}$ to
$4.4\times10^{-3}$). Deletion clears $p<0.001$ and insertion clears $p<0.01$ on
every seed. This is the paper's strongest evidence that the attributions reflect
genuine model behavior rather than post-hoc rationalization.

\paragraph{Caveat.} The effect is robust in aggregate but heterogeneous across
series: the per-series standard deviation is comparable to the mean for deletion
($0.24$ vs.\ mean $0.22$) and appreciably larger than the mean for
insertion ($0.44$ vs.\ mean $0.27$). This spread is why insertion,
despite a \emph{larger} mean gap than deletion, yields a \emph{higher} p-value:
the per-series effect is stronger on average but far more variable. The aggregate
significance is therefore not driven by uniformly large per-series effects, and
individual series can depart substantially from the average. This is the right
claim to make: faithfulness here is a property we establish for the method
\emph{across a population} of series, and clearing significance at $n=30$
\emph{despite} that per-series spread means the effect is large relative to the
variability---strong evidence for the population-level claim, not weak evidence.
What the spread does not license is the stronger reading that every individual
series is explained equally well; we flag it precisely so that reading is not
inferred.

\section{Attribution Analysis}
\label{sec:attribution}
\subsection{Heterogeneity of promotion reliance}
\begin{figure}[t]
\centering
\includegraphics[width=0.72\textwidth]{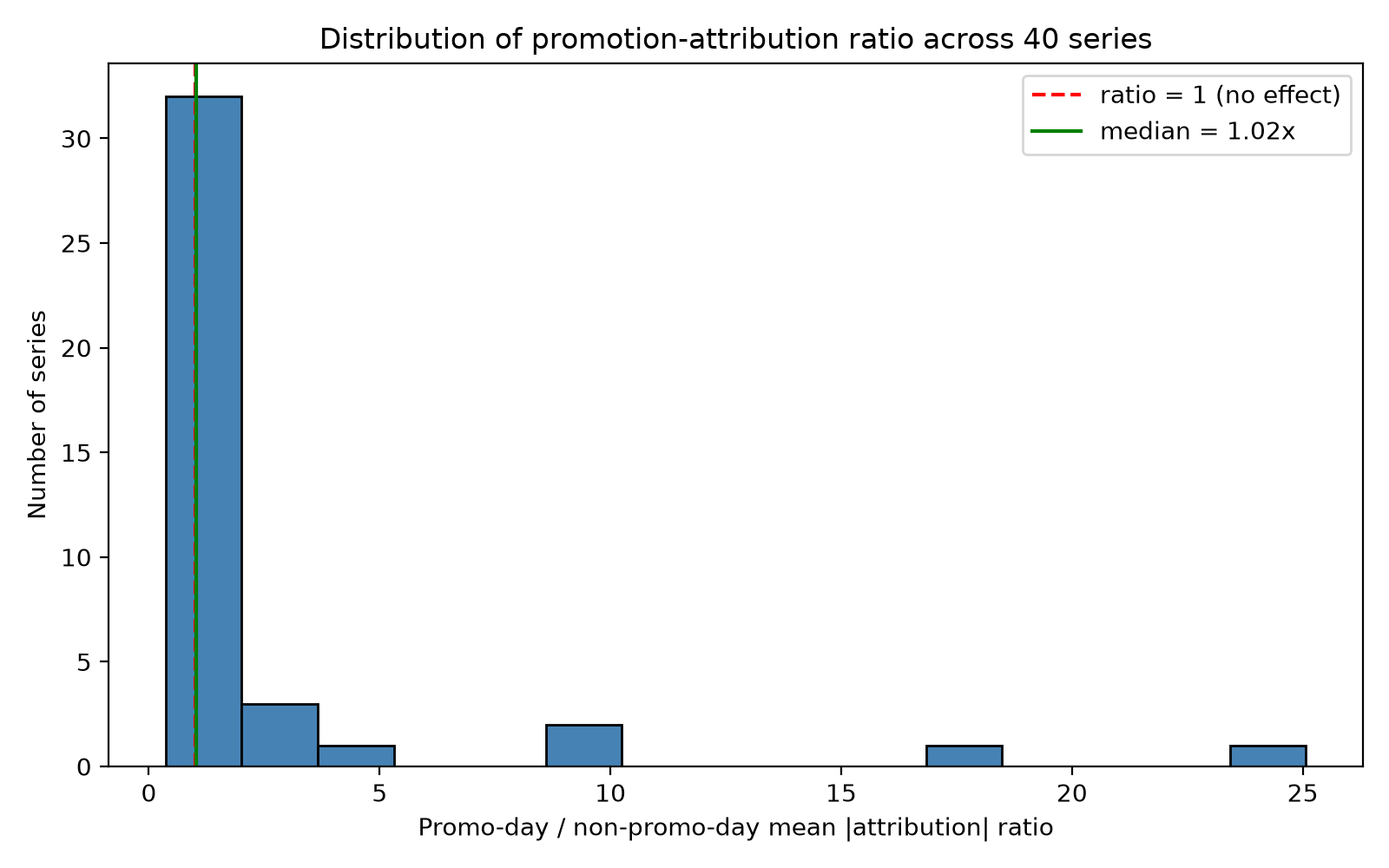}
\caption{Distribution across $40$ series of the ratio between mean absolute
attribution on promotion days and on non-promotion days. The median (green) is
$1.02$; the distribution has a long right tail, and only about $20\%$ of series
exceed a ratio of $2$.}
\label{fig:ratio}
\end{figure}

Faithful attributions let us ask a substantive question: how much does the model
rely on the promotion signal? For each of $40$ series we compute the ratio of
mean absolute attribution on promotion days to that on non-promotion days. We
restrict to the subpopulation of series whose promotion frequency \emph{within
the explained window} lies between $15\%$ and $85\%$ of days, and draw $40$ of
them uniformly at random (fixed seed). The restriction is necessary because the
promotion-day/non-promotion-day ratio is undefined or unstable for series that
are almost never or almost always promoted; sampling uniformly within the
subpopulation, rather than preferring heavily promoted series, avoids biasing the
estimate upward. Figure~\ref{fig:ratio} shows the
distribution. The median ratio is $1.02$---essentially no difference---while the
mean is $2.70$, inflated by a small number of high-ratio outliers. Only about
$20\%$ of series show a ratio above $2$. If anything this makes the finding
conservative: even among series with substantial promotion activity, where a
promotion effect is most likely, the median reliance ratio is $\approx 1.0$.

The honest reading is that promotion reliance is \emph{heterogeneous}: for a
minority of series the model attends strongly to promotions, but for the
majority the promotion signal contributes little to the forecast. This is not
the tidy story that attribution always surfaces promotions, and we present it as
a strength---the method distinguishes the series where a signal matters from
those where it does not, rather than manufacturing an effect everywhere.

\subsection{Weekly cycle}
\begin{figure}[t]
\centering
\includegraphics[width=0.82\textwidth]{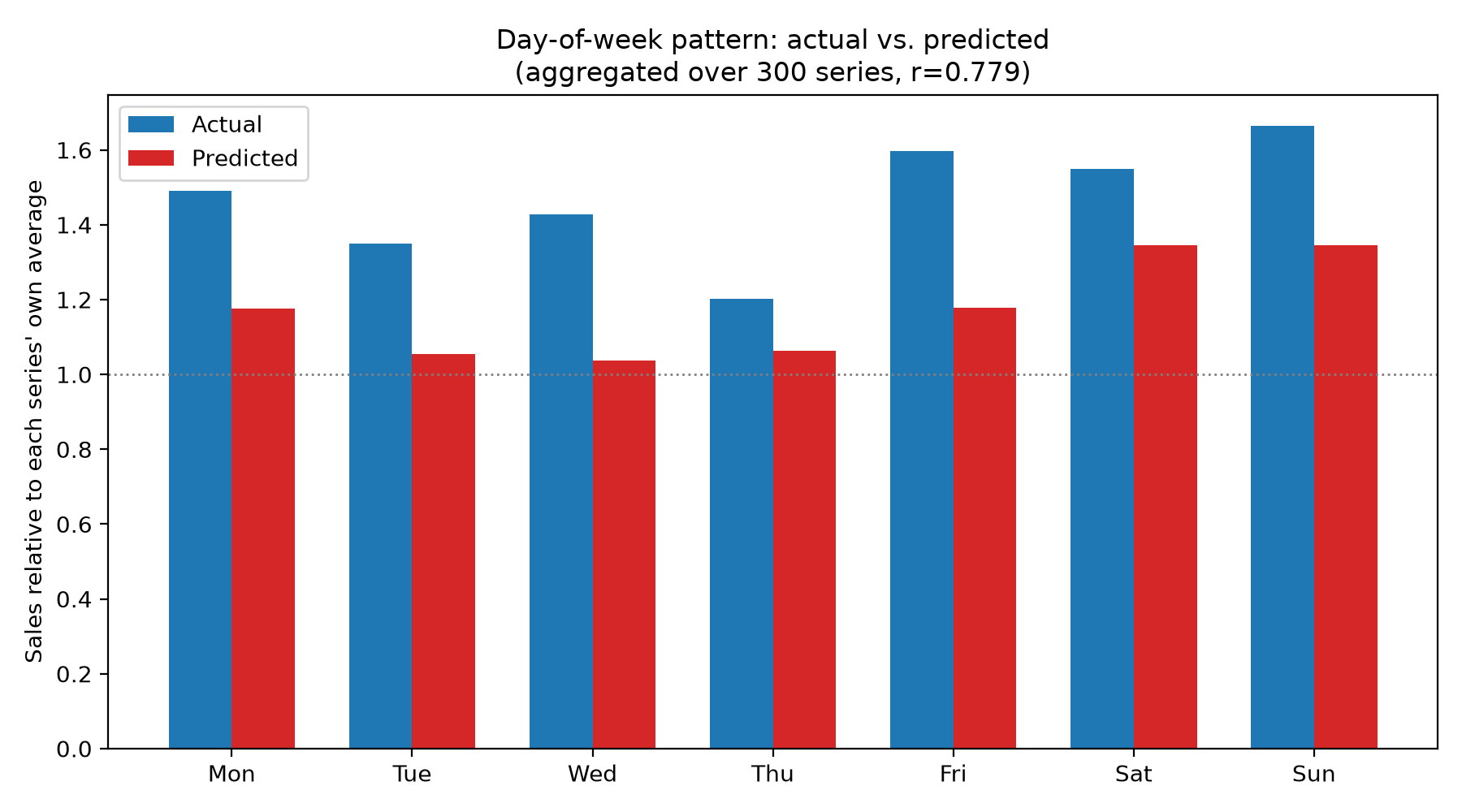}
\caption{Actual vs.\ predicted day-of-week sales profile, aggregated over $300$
series and normalized to each series' own average ($r = 0.78$). The model
captures the shape of the weekly cycle but under-predicts its amplitude: every
predicted bar is closer to $1.0$ than the corresponding actual bar.}
\label{fig:weekly}
\end{figure}

Grocery sales have a strong day-of-week structure. Because only $16$ days are
held out per series---roughly two to three instances of each weekday---we
aggregate over $300$ series to estimate the pattern reliably.
Figure~\ref{fig:weekly} compares the actual and predicted day-of-week profiles,
each normalized to the series' own average. The actual profile shows weekend
peaks and a Thursday low; the model recovers this broad shape ($r = 0.78$,
$p = 0.04$) but damps its amplitude---every predicted day sits closer to the
series average than the corresponding actual value---and places its own trough
midweek rather than on Thursday.

We attribute this damping to the training objective, which optimizes average
forecast accuracy rather than the fidelity of periodic structure; regression
toward the mean is the low-risk solution under such a loss. A related limitation
is that holidays, which drive some of the largest short-term deviations, are
structurally uncapturable with a $90$-day input window---consistent with the
finding of \citet{kechyn2018sales}.

\section{Limitations}
\label{sec:limitations}
\paragraph{Attribution is informative for only a minority of series.} Our
heterogeneity analysis (Section~\ref{sec:attribution}) is also a limitation: for
most series the promotion signal contributes little, so a promotion-focused
explanation is uninformative there. The method reliably distinguishes these
cases rather than inventing a driver, but it cannot surface an actionable factor
where the model does not use one.

\paragraph{Amplitude damping and holidays.} The model recovers the shape of the
weekly cycle but under-predicts its amplitude, and it cannot represent holiday
effects whose spacing exceeds the $90$-day input window---a structural limit
consistent with \citet{kechyn2018sales}. Explanations inherit these blind spots:
the model cannot attribute a forecast to a holiday it never sees.

\paragraph{Order and baseline dependence of contributions.} The exact-sum
guarantee holds for any reveal order and any baseline, but the credit assigned to
an \emph{individual} group is not invariant to either. Because the model is
nonlinear, revealing groups in a different order, or choosing a different neutral
baseline for a withheld input, can shift how much each group receives (though not
the total). We fix one sensible order and a single baseline and report
contributions under that choice; we do not average over orderings as a
Shapley-style method would, trading that invariance for an exact,
directly-measured decomposition.

\paragraph{Attribution concerns the model, not the world.} As stressed in
Section~\ref{sec:method}, a large attribution to the promotion channel means the
model relies on that channel, not that promotions cause sales. Promotions may
coincide with other demand drivers that the channel partly absorbs;
disentangling them is a question of causal inference on the data, which we do not
attempt.

\paragraph{Single dataset and metric.} All results are on one dataset
(Corporaci\'on Favorita) under one metric (NWRMSLE). Whether the faithfulness
result generalizes to other domains and forecasters is open; a replication on a
second domain is the most direct test and a natural next step.

\section{Conclusion}
\label{sec:conclusion}
We added a post-hoc, architecture-agnostic counterfactual attribution layer to a
multi-series WaveNet sales forecaster and asked the question our title poses: how
faithful is the resulting attribution? Deletion and insertion tests answer it
affirmatively and with high significance---the attributions reflect genuine model
behavior---while a heterogeneity analysis shows, just as candidly, that they are
informative for some series and uninformative for others. The method decomposes
any forecast into contributions that sum exactly to the predicted value, and
because it treats the forecaster as a black box, it can explain a model that is
already trained and deployed rather than requiring a switch to an
interpretable-by-design architecture.

We have deliberately not competed on accuracy. The value we claim is narrower
and, we believe, more useful in practice: when a forecast is about to drive an
ordering decision, a faithful account of what drove that number---together with
an honest signal of when such an account is and is not informative---is what lets
a practitioner decide how far to trust it. Extending the evaluation to further
datasets and forecasters, and incorporating known-future covariates, are the
natural next steps.

\subsubsection*{Broader Impact Statement}
This work makes an existing sales forecaster more interpretable; it does not
introduce new forecasting capability or data collection. The main
consideration is that a faithful-looking attribution can still be misread as a
causal claim about the world (e.g.\ that a promotion \emph{caused} sales rather
than that the model \emph{relied} on the promotion signal). We state this
distinction explicitly (Sections~\ref{sec:method} and~\ref{sec:limitations}) and
report where attributions are and are not informative, precisely to discourage
over-trust in automated explanations of consequential decisions.
\bibliography{references}
\bibliographystyle{tmlr}

\end{document}